\documentclass{article}

\usepackage[nonatbib]{arxiv}

\usepackage{amsfonts}       
\usepackage{amsmath,amssymb}
\usepackage{booktabs}       
\usepackage{color}
\usepackage{enumitem}
\usepackage{epsfig}
\usepackage{graphicx}
\usepackage{lineno}
\usepackage[linesnumbered,ruled,vlined]{algorithm2e}
\usepackage{lipsum}
\usepackage{mathrsfs}
\usepackage{microtype}      
\usepackage{multirow}
\usepackage{nicefrac}       
\usepackage{rotating}
\usepackage[T1]{fontenc}    
\usepackage{url}            
\usepackage[utf8]{inputenc} 

\title{Cluster Size Management in Multi-Stage Agglomerative Hierarchical Clustering of Acoustic Speech Segments}

\author{
  Lerato Lerato, Thomas ~Niesler \\
  Department of Electrical and Electronic Engineering\\
  University of Stellenbosch\\
  Stellenbosch,  South Africa \\
  \texttt{llerato@sun.ac.ac.za},  \texttt{trn@sun.ac.za} \\
}

\begin{document}
\maketitle

\begin{abstract}
Agglomerative hierarchical clustering (AHC) requires only the similarity between objects to be known.
This is attractive when clustering signals of varying length,  such as speech, which are not readily represented in fixed-dimensional vector space.
However, AHC is characterised by $O(N^2)$ space and time complexity, making it infeasible for partitioning large datasets. 
This has recently been addressed by an approach based on the iterative re-clustering of independent subsets of the larger dataset.
We show that, due to its iterative nature, this procedure can sometimes lead to unchecked growth of individual subsets, thereby compromising its effectiveness.
We propose the integration of a simple space management strategy into the iterative process, and show experimentally that this leads to no loss in performance in terms of F-measure while guaranteeing that a threshold space complexity is not breached.
\end{abstract}


\section{Introduction}
\label{sec:intro}
Current automatic speech recognition (ASR) systems model words as sequences of sub-word units that generally correspond to the phones of the language.
The decompositions of the words into phones represent the pronunciation variants of the words, and these are usually enumerated in a resource referred to as the pronunciation dictionary.
The definitions of the phones themselves are informed by linguistic expertise, and the pronunciations are often crafted manually by experts.
This process is both time consuming and expensive, and may  not be practical for languages in which the necessary expertise and infrastructure is not available.
Such languages are commonly referred to as under-resourced, and are for example especially prevalent in Africa and parts of Asia \cite{underResource2014}.

For an under-resourced language it would therefore be an advantage if the sub-word units could be determined automatically, without requiring the careful attention of linguistic experts.
A direct approach to this goal is the application of clustering algorithms to unlabelled speech.
Here the objective is to seek out and cluster segments of speech that are similar.
The resulting clusters can then serve as sub-word units, in terms of which words can be modelled in an ASR system.
This segment-and-cluster approach has recently attracted some research attention \cite{LivescusummaryofMyWork,Singh,Wang2015ASWU,Wang2013Posteriorgrams,Bacchiani}.

Since the speech segments that must be clustered are of varying length, they cannot easily be represented in a fixed-dimensional vector space.
However, the similarity between segments can be computed, for example by means of the well-established method of dynamic time warping (DTW).
Given these similarities, agglomerative hierarchical clustering (AHC) can be applied.
However, the space complexity of AHC is $\mathcal{O}(N^2)$ and hence cannot be applied to large datasets as are commonly used to develop ASR systems.
We have recently proposed an iterative divide-and-conquer approach that subdivides AHC of a large dataset into many smaller subsets that can be clustered separately and in parallel, and subsequently recombined \cite{LeratoPLos}.
In many cases this algorithm is able to effectively allow agglomerative hierarchical clustering of datasets that are too large for a straightforward application of the clustering strategy, with no loss in clustering performance.
However, it has since been observed that in some cases individual clusters grow during the iterations and eventually dominate, thereby exceeding available memory and strongly slowing the clustering process.
In this paper we propose an extension to our original algorithm that can be used to remedy this.
By monitoring the occupancy of the clusters in each subset and iteratively subdividing them when a threshold size is exceeded, maximum memory usage can be guaranteed.
This in principle makes the clustering of very large databases possible given available memory that is much too small for full AHC.

After presenting a review of the literature, we summarise our previously proposed algorithm.
We then present an extended algorithm that enforces a hard limit on the size any cluster may take on in the sub-stages.
Finally, we apply this algorithm to a number of speech datasets and demonstrated its effectiveness.

\section{Related Work}
\label{sec:literature} 

Wang \emph{et al} \cite{Wang2015ASWU} have considered the clustering of speech segments using spectral clustering \cite{spectral_von07tut, spectralLandmarkChen11}. 
The speech signal is first divided into non-overlapping segments using an Euclidean-based distortion measure.
Using hierarchical clustering, the authors determine the number of segments and their boundaries. 
The dataset is then represented in terms of a distance matrix whose rows represent number of Gaussians while the columns correspond to the number of segments.
Gaussian component clustering (GCC) and segment clustering (SC) are applied, where GCC applies spectral clustering to a set of Gaussian components and SC applies spectral clustering to a large number of speech segments.
The final step employs multiview segment clustering (MSC), which takes a linear combination of the Laplacian matrices obtained from different posterior representations and derives a single spectral embedding representation for each segment. 
The OGI-MT2012 corpora were used for experimentation. 
Clusterings were evaluated using both purity and normalised mutual information (NMI) \cite{NMIVinhinformationtheoretic}.
The authors had previously reported similar work also utilising GCC and SC where data was converted into segment-level Gaussian posteriograms (SGP's) and then consolidated into distance matrix of size \emph{M} Gaussians by \emph{N} segments \cite{Wang2013Posteriorgrams}. 
In this case clustering is carried out using the normalized cut  \cite{Shi97normalizedcutsShi97} approach with a pre-determined number of clusters. 

Bacchiani and Ostendorf \cite{Bacchiani} implement a segment-and-cluster approach to jointly learn a unit inventory and corresponding lexicon from data \cite{Bacchiani_ICASSP96SegClust, Bacchiani}. 
Unconstrained acoustic segmentation is applied to utterances with known word boundaries. 
Each resulting acoustic segment is assigned to the word token with which it overlaps the most, using a context-dependent HMM system. 
Given the pronunciation lengths, a constrained acoustic segmentation is applied by using dynamic programming with the assumption that each word boundary coincides with an acoustic segment boundary. 
The inventory which contains an equal number of segments for each training token of a particular word is generated. 
Divisive and k-means clustering is applied to the speech segments, here the number of clusters is heuristically determined. 
The resulting clusters are considered to be acoustic subword units and are refined using Viterbi training.  

The clustering of acoustic speech segments was also investigated by Paliwal \cite{PaliwalASWU1990}. 
Here each segment is first represented by its centroid.
Next, the centroids are clustered using k-means, assuming a known number of clusters.  
A maximum likelihood segmentation process produces a codebook containing $N$ entries, each defining one acoustic sub-word unit cluster \cite{SvendsenICASSP1989}.
A similar process is reported by Lee \emph{et al} in \cite{LeeAndSoongICASSP88}. 

Agglomerative hierarchical clustering of acoustic speech segments is not popular in the literature. One example of AHC application to acoustic segments is the work of Mak and Barnard \cite{Mak96phoneclustering} where clustering of biphones is carried out using the Bhattacharyya distance. Although this distance is probabilistically measured, the singleton biphones at the top of the dendrogram are each represented by a Gaussian acoustic model. Agglomerative hierarchical clustering (AHC) is used to merge similar biphones using the Bhattacharyya distance until only one cluster is left. Building Gaussian models for a single biphone leads to a possibility of insufficient data and incomplete biphone coverage. This  is solved by  a proposed two-level clustering algorithm. The  first step is to cluster monophones using conventional AHC until a fair amount of data enough to create a model is obtained. Acoustic models are then re-computed, after which a final AHC  step is performed. The OGI\_TS corpus is used to evaluate the results.

\section{Agglomerative Hierarchical Clustering of Speech Segments}
\label{sec:AHCdescription}

Consider a dataset $\mathscr{X}$ consisting of $N$ objects as indicated in Equation~\ref{eqn:genericdata}.

\begin{equation}\label{eqn:genericdata}
\mathscr{X}=\lbrace \mathbf{X}_1, \mathbf{X}_2, \mathbf{X}_3 ,..., \mathbf{X}_N \rbrace
\end{equation}

The objective of the clustering operation is to partition $\mathscr{X}$ into a set of $K$ clusters $\mathscr{C} = \{ \mathbf{C}_1, \mathbf{C}_2, ...,\mathbf{C}_K \}$ while ensuring that each cluster $\mathbf{C}_k$ is populated by similar objects.
Furthermore,  $\mathbf{C}_p \cap \mathbf{C}_q = \emptyset$ for $p,q=1,2,...,K$ where $p \neq q$, while $ \mathbf{C}_1 \cup \mathbf{C}_2 \cup ...\cup \ \mathbf{C}_K = \mathscr{X}$. 

In most cluster analysis literature, each object is a point in a $d$-dimensional space  \cite{JainBook1988,IRBook,zhang1996birch}.  
We will however consider the situation where each object consists of a time series of $n$ $d$-dimensional points, so that $\mathbf{X}_i =\{ \mathbf{x}_{i1}, \mathbf{x}_{i2}, \mathbf{x}_{i3} ,..., \mathbf{x}_{in} \}$ with $\mathbf{x}_{ij} \in \mathbb{R}^d$.
The similarity between two objects $\mathbf{X}_i$ and $\mathbf{X}_j$ is given by $D(\mathbf{X}_i,\mathbf{X}_j)$.

As input, agglomerative hierarchical clustering (AHC) takes a lower or upper triangular similarity matrix $D$ whose entries are given by all possible pairwise distances $D(\mathbf{X}_i,\mathbf{X}_j) \,\forall\, i,j$ and where it has been assumed that $D(\mathbf{X}_i,\mathbf{X}_j) = D(\mathbf{X}_j,\mathbf{X}_i)$.  
In our case, these distances are computed using the dynamic time warping (DTW) algorithm, which is a popular similarity measure used in speech processing \cite{DTW_speech_Myers,Yu2007CTS}. 
DTW recursively determines the best alignment between the two segments by minimizing a cumulative cost that is commonly based on the Euclidean distances between time aligned time-series vectors. 

AHC begins with the assumption that each object is the sole occupant of its own cluster. 
A binary tree structure referred to as a \emph{dendrogram} is created by successively merging the closest cluster pairs until a single cluster remains \cite{murtagh2011methods}.
Inter-cluster distances are normally calculated using linkage methods \cite{IRBook}. 
Popular choices are single, average and complete linkage. 
We have chosen the Ward method \cite{Ward1963,murtaghWard2014} to calculate inter-cluster similarity because it has been found by other researchers to outperform other linkage methods \cite{JainBook1988} and because it has lead to consistently better results in our previous experiments.  
The Ward method is a minimum variance criterion that minimises the total within-cluster variance.  
It uses the squared distance between cluster centres or the Euclidean distance between the individual objects \cite{murtagh2011methods,xuClustering2005survey}.
 
\section{Multi-Stage Agglomerative Hierarchical Clustering}
\label{sec:mahc}

Agglomerative hierarchical clustering as described in the previous section requires the computation and storage of $M=N(N-1)/2$ similarities.
This  $O(N^2)$ space and computational complexity becomes impractical for large $N$. 
In previous work~\cite{LeratoPLos}, we have proposed a solution in the form of a divide-and-conquer strategy which we termed multi-stage AHC (MAHC).

In a first stage, this algorithm divides the dataset $\mathscr{X}$ into $P$ subsets, where $P$ is heuristically determined by the available memory but also by the available number of processors.
AHC is applied to each of these $P$ subsets, either sequentially or in parallel, to generate $P$ sets of clusters $\mathscr{C}_1, \mathscr{C}_2,  \ldots  \mathscr{C}_P$ and $\mathscr{C}_p = \{ \mathbf{C}_{p1}, \mathbf{C}_{p2}, ...,\mathbf{C}_{pK_p} \}$.
For each subset, the number of clusters $K_p$ is automatically determined using the L method as implemented by Salvador and Chan \cite{Salvador2004Knee}.

In a second stage, a medoid $\mathbf{\bar{X}}_p$ is computed for each of the $K_p$ clusters in each of the $P$ subsets.
The resulting set of medoids $\{ \mathbf{\bar{X}}_1, \mathbf{\bar{X}}_2,...,\mathbf{\bar{X}}_S \}$ where $S = K_1 + K_2 +...+ K_P$ is then itself grouped into $P$ clusters, again using AHC. 
Finally, these $P$ clusters are used to repartition $\mathscr{X}$ into $P$ new subsets, thereby re-initialising the MAHC process and rendering it iterative.

We were able to show that, in terms of F-measure, MAHC matches or even surpasses conventional AHC within a small number of iterations. 
It was also demonstrated that, in terms of F-measure, MAHC provided an improvement over parallel spectral clustering \cite{chen2011parallel}.
Ideally, MAHC exhibits $O(\frac{N^2}{P^2})$ space and computational complexity for each subset, and during practical evaluation this was usually observed to be the case.
However, due the iterative nature of the algorithm, it is possible that one or more of the $P$ subsets grows to contain substantially more than $\frac{N}{P}$ objects. 
These oversized clusters then dominate the computational capacity and storage requirements of MAHC, whose complexity can in the worst case again approach $O(N^2)$. 

\begin{figure}[!h]
	\centering
	\includegraphics[scale=0.6]{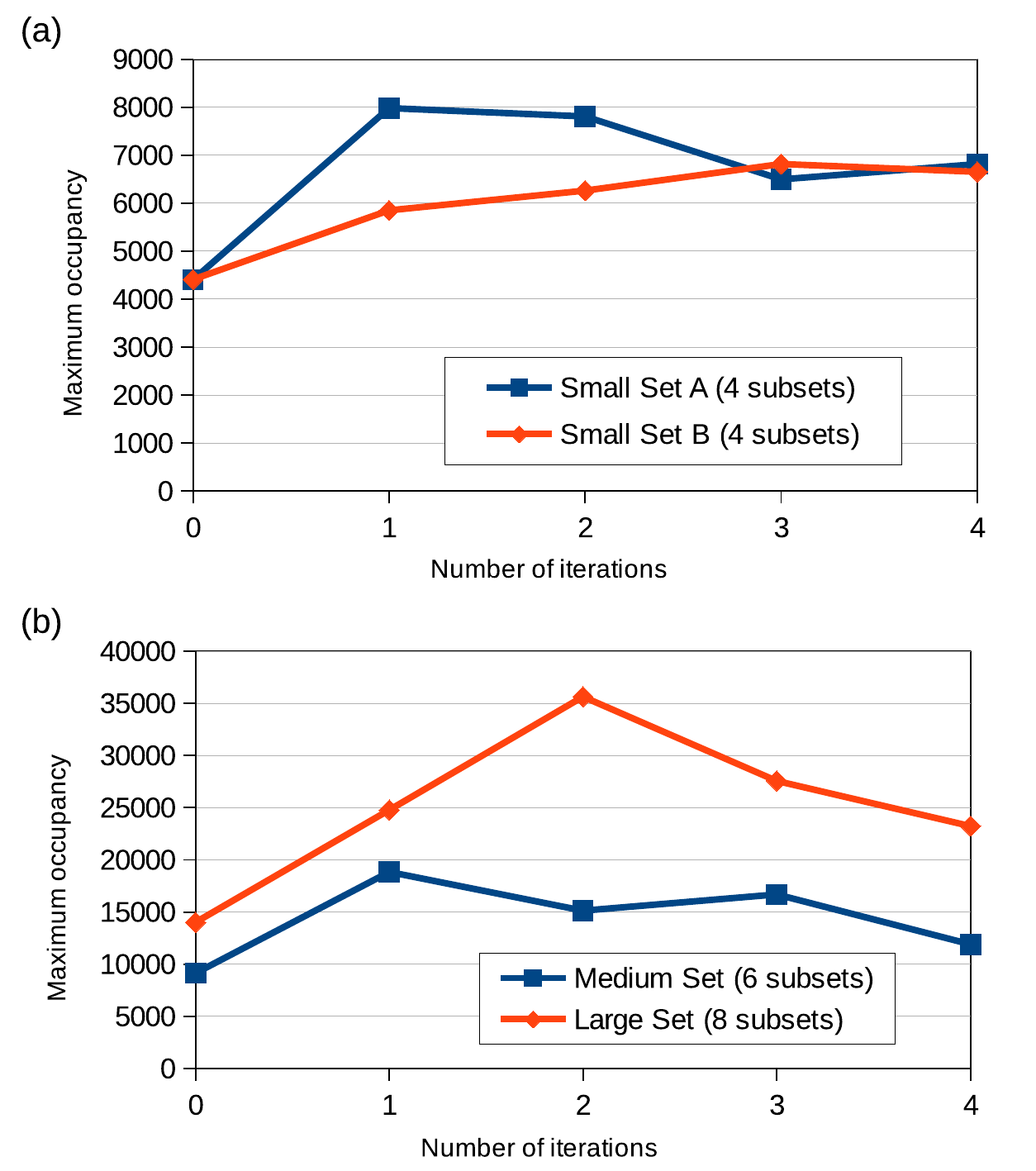}
	\caption{Total membership per iteration of the subset containing the largest number of speech segments when applying MAHC to (a) Small Set A and Small Set B in both cases with $P=4$ subsets and (b) the Medium Set with $P=6$ subsets and the Large Set with $P=8$ subsets.}
	\label{fig:mahcOcc}
\end{figure}

Figure~\ref{fig:mahcOcc} illustrates how the number of occupants of the largest subset evolves during 5 iterations of the MAHC algorithm in example applications to the datasets described later in Section~\ref{sec:data}.
At iteration 0, the occupancy corresponds to evenly-divided subsets, i.e. $\frac{P}{N}$. 
For all four datasets, the occupancy of the largest cluster grows during at least the first two iterations.
This increase can be mild, as it is for the Medium Set in  Figure~\ref{fig:mahcOcc} (b).
However for Small Set A in Figure~\ref{fig:mahcOcc} (a) as well as for the Large Set  in  Figure~\ref{fig:mahcOcc} (b) the occupancy of the largest cluster grows to approximately twice its initial value at some point in the iterative process.

A chief objective of the MAHC algorithm was to ensure that the similarity matrices that must be computed remain manageable in size, so that they can be stored in memory and do not have to be relegated to disk, for example.
However Figure~\ref{fig:mahcOcc} shows that the occupancy of individual subsets may grow substantially.
In particular there is no guarantee that the practically available memory will not be exceeded.

\section{Cluster Size Management for Multi-Stage Agglomerative Hierarchical Clustering}
\label{sec:MAHC+M}

\begin{figure}[!ht]
\centering
\includegraphics[scale=0.22]{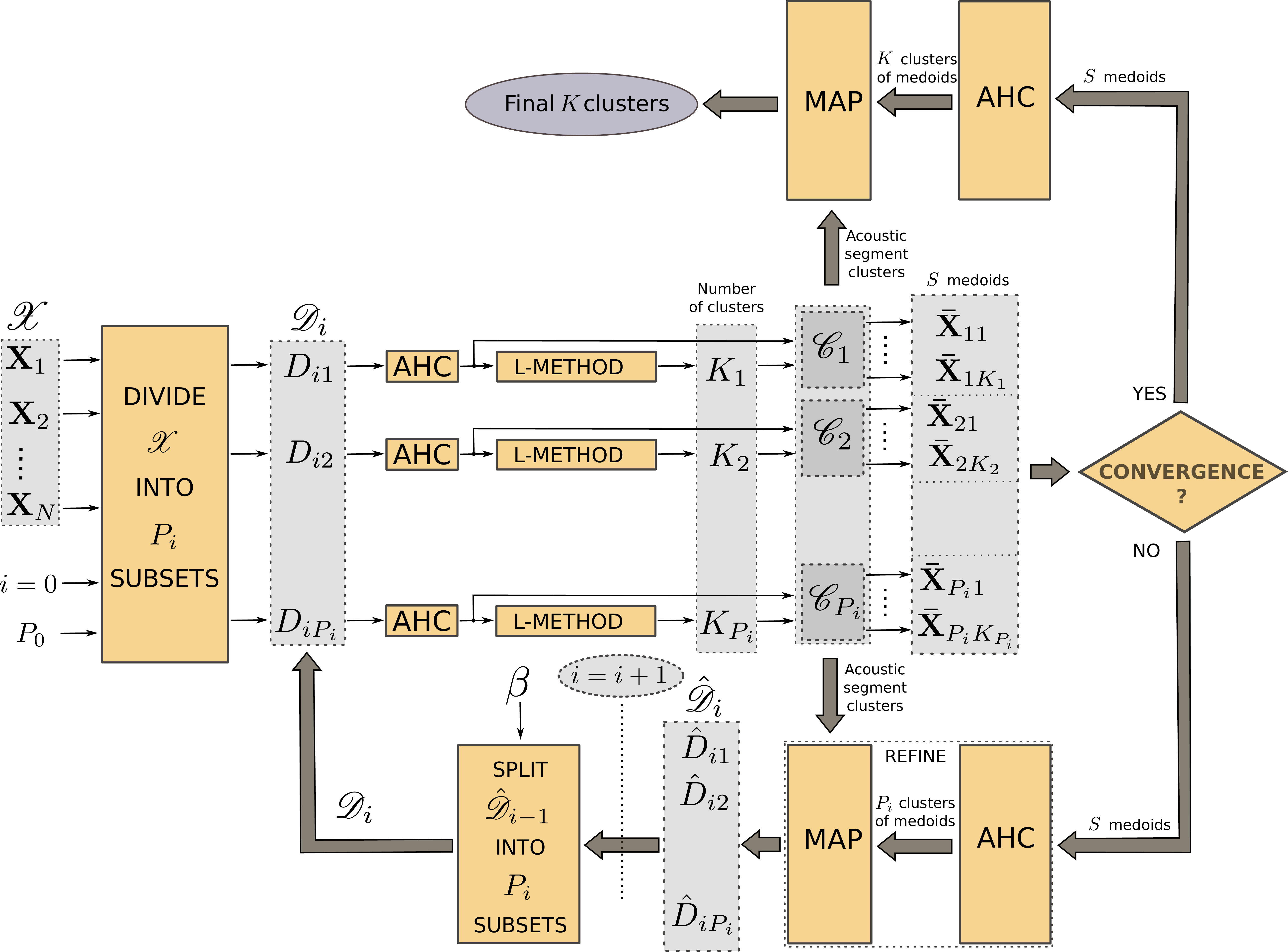}
\caption{Modified agglomerative hierarchical clustering (MAHC) with cluster size management, as also described in Algorithm~\ref{algo:MAHC+M}.} 
\label{fig:MAHC+M}
\end{figure}

We will address the runaway growth in occupancy of certain clusters during MAHC by repeatedly subdividing the offending clusters at each iteration of the algorithm.
We will also consider the appropriateness of merging clusters when they become too small.
This seeks to maintain the advantages offered by MAHC while guaranteeing that no subset grows too large for the available computational and storage resources. 
It requires the number of subsets $P$ to be allowed to vary between iterations, as illustrated in Figure~\ref{fig:MAHC+M}.
 
The parameters of the algorithm are:
\begin{enumerate}
\item The  initial number of subsets, $P_0$ . 
\item The final number of clusters, \emph{K}.
\item An integer threshold $\beta$ indicating the largest number of objects any subset is allowed contain.
\end{enumerate}

The steps of the clustering procedure are described in Algorithm~\ref{algo:MAHC+M}.
The threshold $\beta$ will usually be dictated directly by available memory and hence the largest cluster for which the distance matrix can be computed. 
A split step in the algorithm uses $\beta$ to subdivide all subsets that have exceeded $\beta$ and ensures that all subsets delivered to the next iteration of the algorithm are within this limit. 
Acoustic data are initially divided into $P_0$ subsets, also in  accordance with available memory and processors.

Convergence can be decided on the basis of a settling in the number of subsets $P_i$, or simply by terminating the clustering procedure after a fixed number of iterations.
For MAHC+M it has previously been empirically been demonstrated that the final number of clusters is well approximated by the total number of clusters resulting from the first stage of the algorithm~\cite{LeratoPLos}.
We have verified that this approximation $K = \sum\limits_{j=1}^{P_i}K_j$ remains valid after the introduction of cluster size management and can therefore again be used to automatically determine a suitable value for the final number of clusters $K$.

\begin{algorithm}[!hb]
\KwIn{$N$ acoustic segments $\mathscr{X}=\lbrace \mathbf{X}_1, \mathbf{X}_2, \mathbf{X}_3 ,..., \mathbf{X}_N \rbrace$; initial number of subsets $P_0$; integer threshold $\beta$; .}
\KwOut{$K$ clusters $\mathscr{C} = \{ \mathbf{C}_1, \mathbf{C}_2, ...,\mathbf{C}_K \}$}
$i=0$ \;
Divide $N$ acoustic segments into $P_i$ subsets $\mathscr{D}_i = \{ \mathbf{D}_{i1}, \mathbf{D}_{i2}, ...,\mathbf{D}_{iP_i} \}$ \;
Independently apply AHC to each subset, resulting in $P_i$ dendrograms \;
Use the L method to determine the optimal number of clusters $K_p$, $p = 1, 2,\ldots P_i$, for each of the $P_i$ dendrograms in Step~3. This results in $P_i$ sets of clusters $\mathscr{C}_p = \{ \mathbf{C}_{p1}, \mathbf{C}_{p2}, ...,\mathbf{C}_{pK_p} \}$ with $p = 1,2,\ldots P_i$\; 
Find the medoid $\mathbf{\bar{X}}_{pk}$ of each cluster $\mathbf{C}_{pk}$ where, $p=1,2,...P_i$ and $k = 1,2,\ldots K_p$. This results in a set of $S$ medoids $\mathscr{\bar{X}}$, where $S = \sum\limits_{p=1}^{P_i} K_p$  \;
If $i>2$ and convergence has been achieved go to Step~11 (conclude) \;
Divide the $S$ medoids obtained in the Step~5 into $P_i$ clusters using AHC \; 
Map the members of each cluster $\mathbf{C}_{pk}$ to one of $P_i$ new subsets  $\hat{\mathscr{D}}_i = \{ \hat{\mathbf{D}}_{i1}, \hat{\mathbf{D}}_{i2}, ...,\hat{\mathbf{D}}_{iP_i} \}$
according to the result of the previous step (refine)\;
Consider each new subset $\hat{\mathbf{D}}_{ij}$ $j = 1,2,\ldots P_i$ and if it contains more than $\beta$ acoustic segments, subdivide it evenly to ensure that the limit $\beta$ is not exceeded (split) \;
Let the total number of subsets resulting from the previous step be $P_{i+1}$ and the subsets themselves be denoted by  $\mathscr{D}_{i+1}$ \;
$i = i + 1$  \;
Go to Step~3 (iterate)\;
Divide the $S$ medoids obtained in the previous step into $K = \sum\limits_{j=1}^{P_i}K_j$ clusters using AHC \; 
Map the members of each cluster $\mathbf{C}_{pk}$ to one of $K$ new subsets according to the result of the previous step \;
The $K$ subsets obtained in the previous step are the final clustering result.
\caption{Modified agglomerative hierarchical clustering (MAHC) with cluster size management, as also described in Figure~\ref{fig:MAHC+M}}
\label{algo:MAHC+M}
\end{algorithm}

\section{Experimental setup}

\subsection{Data}
\label{sec:data}
Data used in all our experiments are acoustic segments taken from a well-established TIMIT speech corpus \cite{Phone_class_Halberstadt,TIMIT_phone_reg}. 
TIMIT contains a total of 6300 sentences recorded from 630 speakers. 
Each speaker reads 10 sentences, the first 2 of which are identical across all speakers in the database. 
To avoid bias, these two sentences have been excluded in all our experiments. 
The TIMIT corpus is chosen because it includes accurate time-aligned phonetic transcriptions meaning that both phonetic labels and their start/end times are provided. 
We will consider triphones \cite{Phone_class_Halberstadt}, which are phones in specific left and right contexts, as our speech segments to be clustered. 
We used a set of 42 base phones in our experiments, corresponding to triphones that are at least 5 milliseconds long. 
Pauses were excluded from our dataset.

From the TIMIT data we have compiled 4 datasets, varying in size.  
Table~\ref{tab:timit} shows the number of segments (objects) in each dataset, as well as the true number of classes, the range of class cardinality and the total number of feature vectors.

\begin{table}[!h]
\centering

\caption{Composition of experimental data, indicating the total number of speech segments, the total number of classes (unique triphones), the frequency of occurrence of each triphone, the total number of feature vectors in $\mathbb{R}^{39}$ and the number of similarities $N(N-1)/2$  which would have to be computed for straightforward application of AHC.}\label{tab:timit}

\begin{tabular}{lrrcrr}
\hline 
Dataset & Segments & Classes & Frequency  & Vectors & Similarities  \\
\hline 
 Small Set A & 17 611 & 280 & 50--373 & 274 677 & $0.16 \times 10^9$
 \\
 
 Small Set B & 17 640 & 636 & 26--49 & 301 026  & $0.16 \times 10^9$
\\ 
 
Medium Set & 54 787 & 1 387 & 20--373 & 910 189 & $1.5 \times 10^9$
\\
 
Large Set & 123 182 & 19 223 & 1--373 & 2 193 793 & $7.6 \times 10^9$
\\ 
\hline 

\end{tabular}

\end{table}

\begin{figure}[!h]
\centering
\includegraphics[scale=0.65]{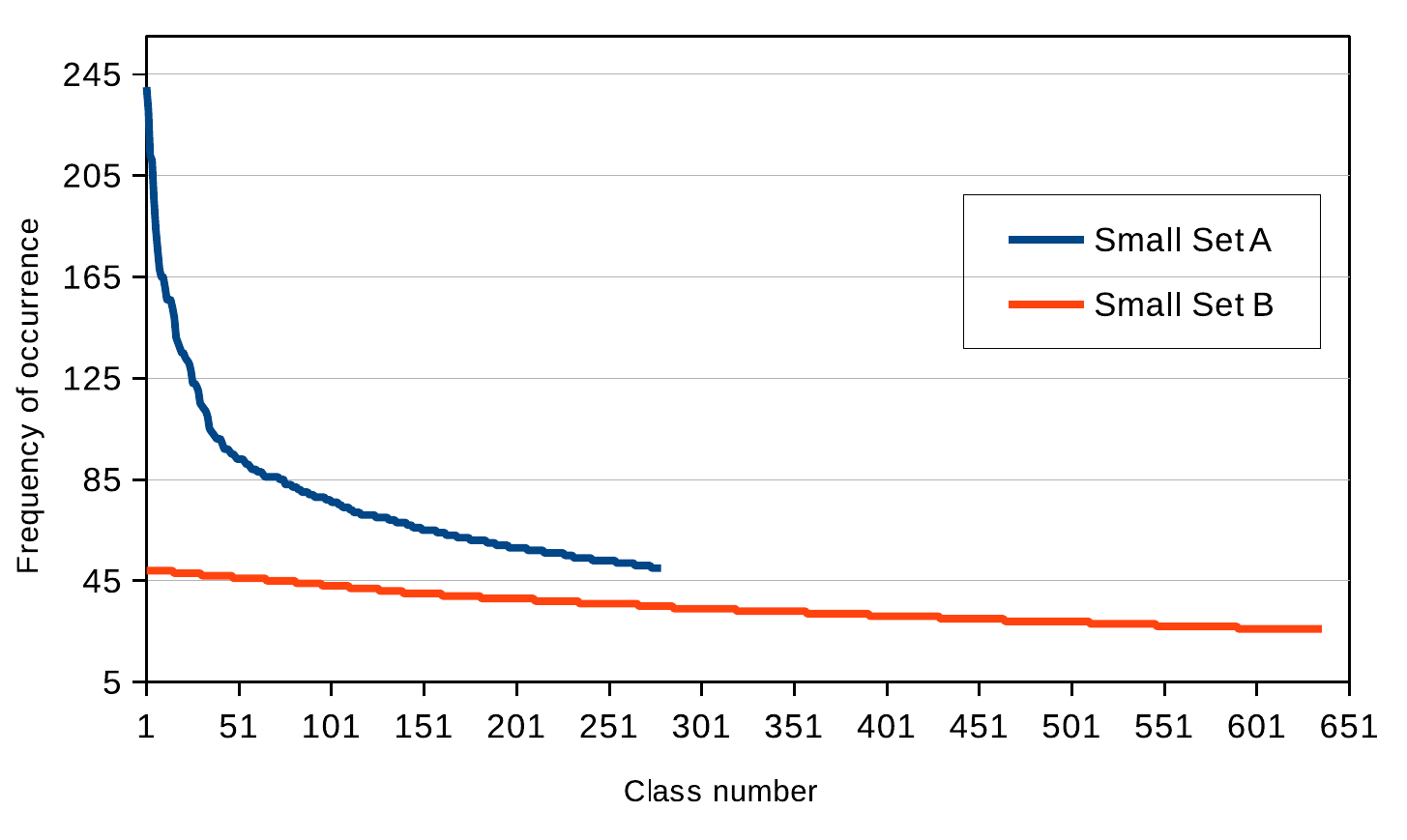}
\caption{Distribution of the number of segments per class for Small Set A and Small Set B.}
\label{fig:ssetassetbdist}
\end{figure}

As depicted in Figure ~\ref{fig:ssetassetbdist}, Small Set A is more skewed than Small Set B and some classes have many more members than others. 
This is also reflected by the frequency of occurrence shown in Table~\ref{tab:timit}. 
The Medium Set and Large Set are skewed in the same way as Small Set A, since this is the type of distribution one may expect in unconstrained speech. 

During data preparation, each acoustic segment is represented as a series of 39 dimensional feature vectors consisting of 12 Mel frequency cepstral coefficients (MFCCs), log frame energy, and their first and second differentials. The MFCCs were chosen on the basis of their well-established popularity in high performance speech processing systems \cite{triphone_clust_Imperl}. Feature vectors are extracted from data frames that are 10ms in length, and consecutive frames overlap by 5ms (50\%). The MFCCs were computed using a HTK \cite{young2002htk}.

%

\subsection{The F-measure}
\label{subsec:fmeasure}

We will use the F-measure to quantify the quality of a division of the acoustic segments in the dataset into one of \emph{K} clusters. 
The F-measure  assumes  that each acoustic segment, $\mathbf{X}_i$, has a known class label representing the ground truth \cite{Fmeasure_Larsen,IRBook,Wu2009Towards}. 
It is based on the measures recall and precision for each cluster with respect to each class in the dataset. 

Assume that for class \emph{l} and cluster \emph{k} we know  $\emph{n}_{kl}$, the number of objects of class \emph{l} that are in cluster \emph{k}; and $\emph{n}_k$ is the total  number of objects in cluster \emph{k} and $\emph{n}_l$ is the number of objects in class \emph{l}. 
The precision and recall are given by Equations \ref{eqn:precision} and \ref{eqn:recall} respectively.

\begin{equation}
pr(k,l)=\dfrac{n_{kl}}{n_k}
\label{eqn:precision}
\end{equation}

\begin{equation}
re(k,l)=\dfrac{n_{kl}}{n_l}
\label{eqn:recall}
\end{equation}

Precision indicates the degree to which a cluster is dominated by a particular class, while recall indicates the degree to which a particular class is concentrated in a specific cluster. 
The F-measure, $F(k,l)$, is calculated as follows:
 
\begin{equation}
F(k,l)=\dfrac{2 \times re(k,l) \times pr(k,l)}{re(k,l) + pr(k,l)}
\label{eqn:Fmeasure}
\end{equation}

where $k=1,2,...,K$ and $l=1,2,...,L$. 
An F-measure $F(k,l)=1$ indicates that each class occurs exclusively in exactly one cluster, which is the perfect clustering result.

\section{Experimental Evaluation }
\label{sec:experiments}

The algorithm described in Section~\ref{sec:MAHC+M} is applied to the datasets described in Section~\ref{sec:data}.
For comparison, results without cluster size management as described in Section~\ref{sec:mahc} are also shown.

\begin{figure}[!h]
\centering
\includegraphics[scale=0.55]{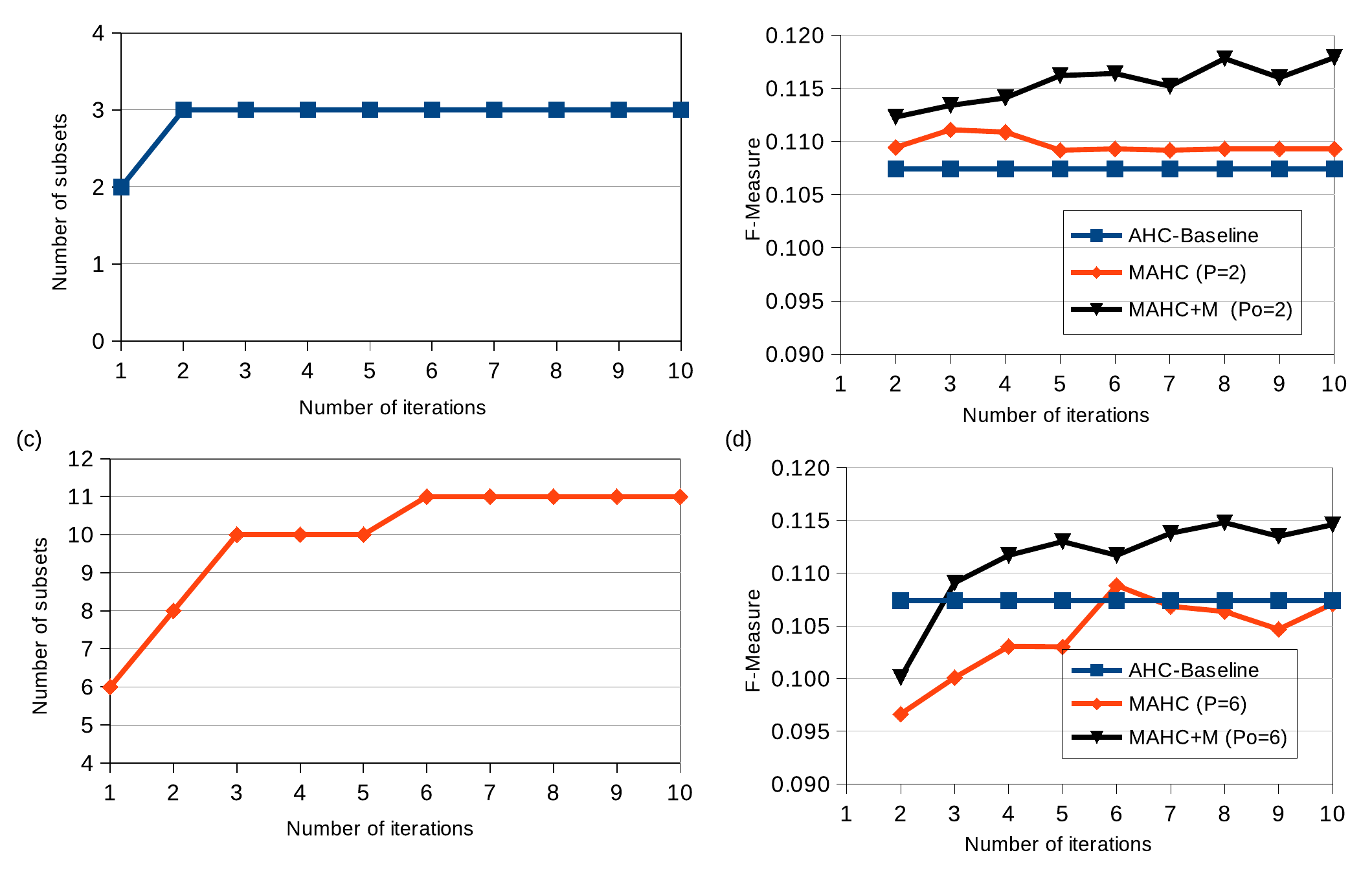}
\caption{Number of subsets  $P_i$ as well as F-measure for each iteration when applying classical agglomerative hierarchical clustering (AHC), modified AHC (MAHC) and MAHC  with cluster size management (MAHC+M) to Small Set A with an initial number of subsets of $P_0=2$ (a and b)  and $P_0=6$ (c and d).}
\label{fig:FigA}
\end{figure}

\begin{figure}[!h]
\centering
\includegraphics[scale=0.55]{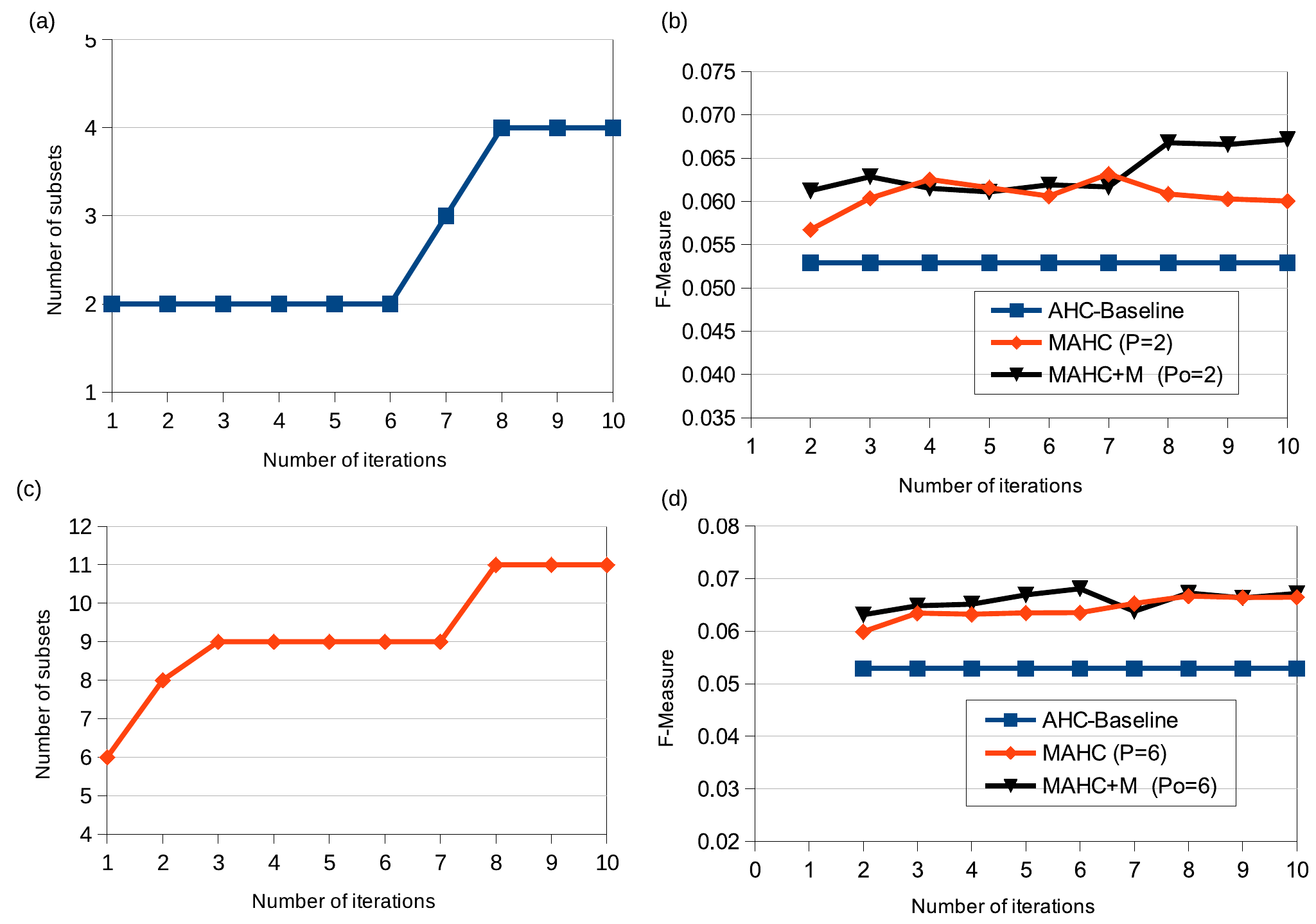}
\caption{Number of subsets  $P_i$ as well as F-measure for each iteration when applying classical agglomerative hierarchical clustering (AHC), modified AHC (MAHC) and MAHC  with cluster size management (MAHC+M) to Small Set B with an initial number of subsets of $P_0=2$ (a and b) and $P_0=6$ (c and d).}
\label{fig:FigB}
\end{figure}

Figures~\ref{fig:FigA} (a) and (c) show the number of subsets per iteration $P_i$ when the Small Set A data are initially divided into $P_0=2$ and $P_0=6$ subsets respectively. 
The corresponding F-measure plots are shown in shown in Figures~\ref{fig:FigA} (b) and (d).  
We see that, for Small Set A, the the introduction of cluster size management has led to some improvement in terms of F-measure for both $P_0=2$ and $P_0=6$. 
Figure~\ref{fig:FigB} shows the results of a corresponding set of experiments for Small Set B, which is similar in size to Small Set A but not as skewed (Figure~\ref{fig:ssetassetbdist}).
We see that also in this case cluster size management has resulted in no deterioration in terms of F-measure.

\begin{figure}[!h]
\centering
\includegraphics[scale=0.5]{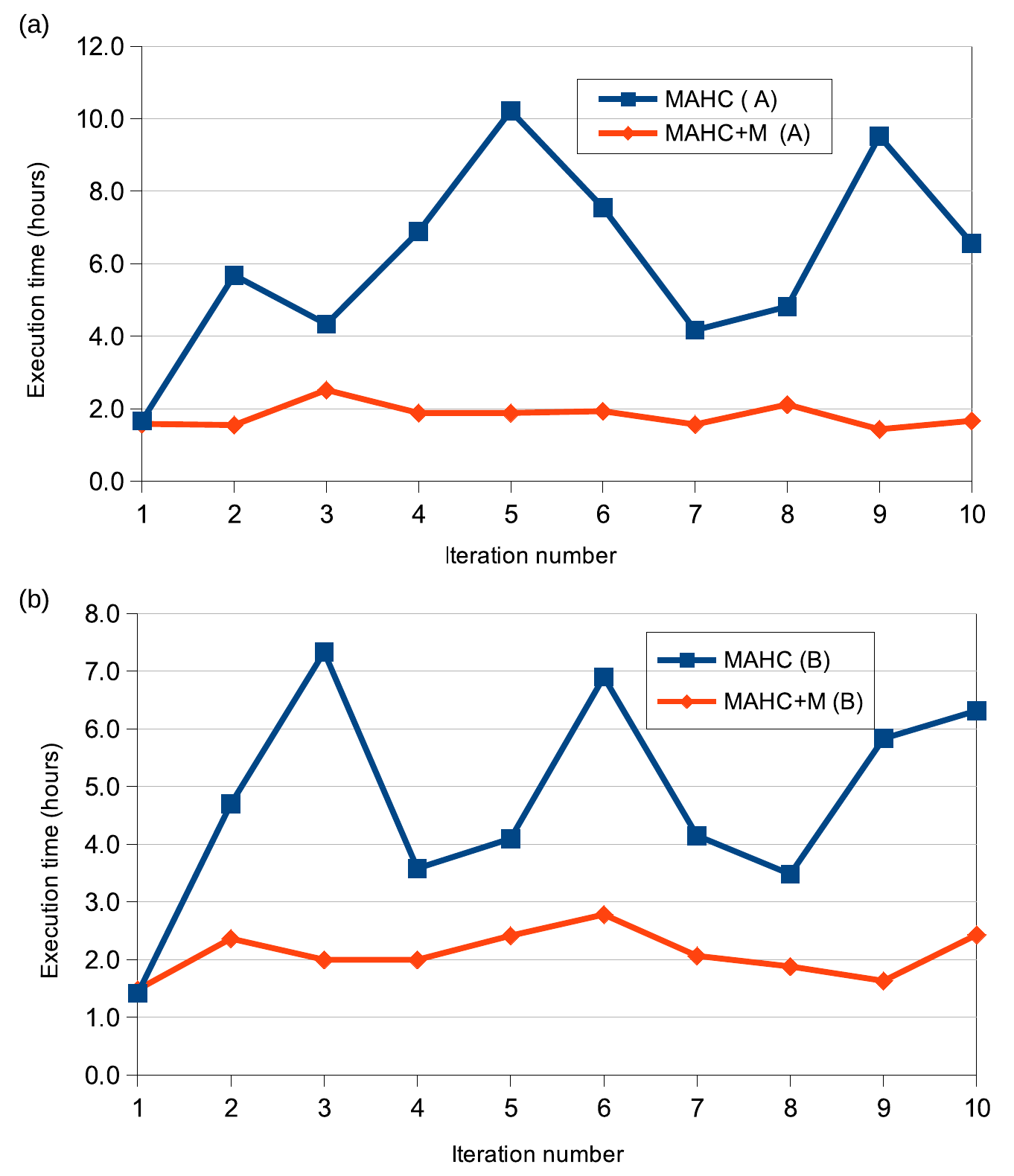}
\caption{Per-iteration execution time of modified agglomerative hierarchical clustering with (MAHC+M) and without (MAHC) cluster size management with $P_0=6$ initial subsets for (a) Small Set A and (b) Small Set B. }
\label{fig:timeAB}
\end{figure}

To obtain an indication of the practical impact on processing time afforded by the introduction of cluster size management, Figure~\ref{fig:timeAB} shows the measured time (in hours) taken per iteration to cluster Small Set A and Small Set B with $P_0=6$.
These small datasets were chosen because they allow clustering to be performed on a normal stand-alone workstation, in our case an Intel Core i7 with four cores, running at 3.40Ghz and with 32GB of RAM.
Figure~\ref{fig:timeAB} indicates a reduction in processing time of up to a factor of five, while Figures~\ref{fig:FigA}~(d) and~\ref{fig:FigB}~(d) have already indicated that this savings does not incur a penalty in terms of F-measure. 

\begin{figure}[!h]
\centering
\includegraphics[scale=0.5]{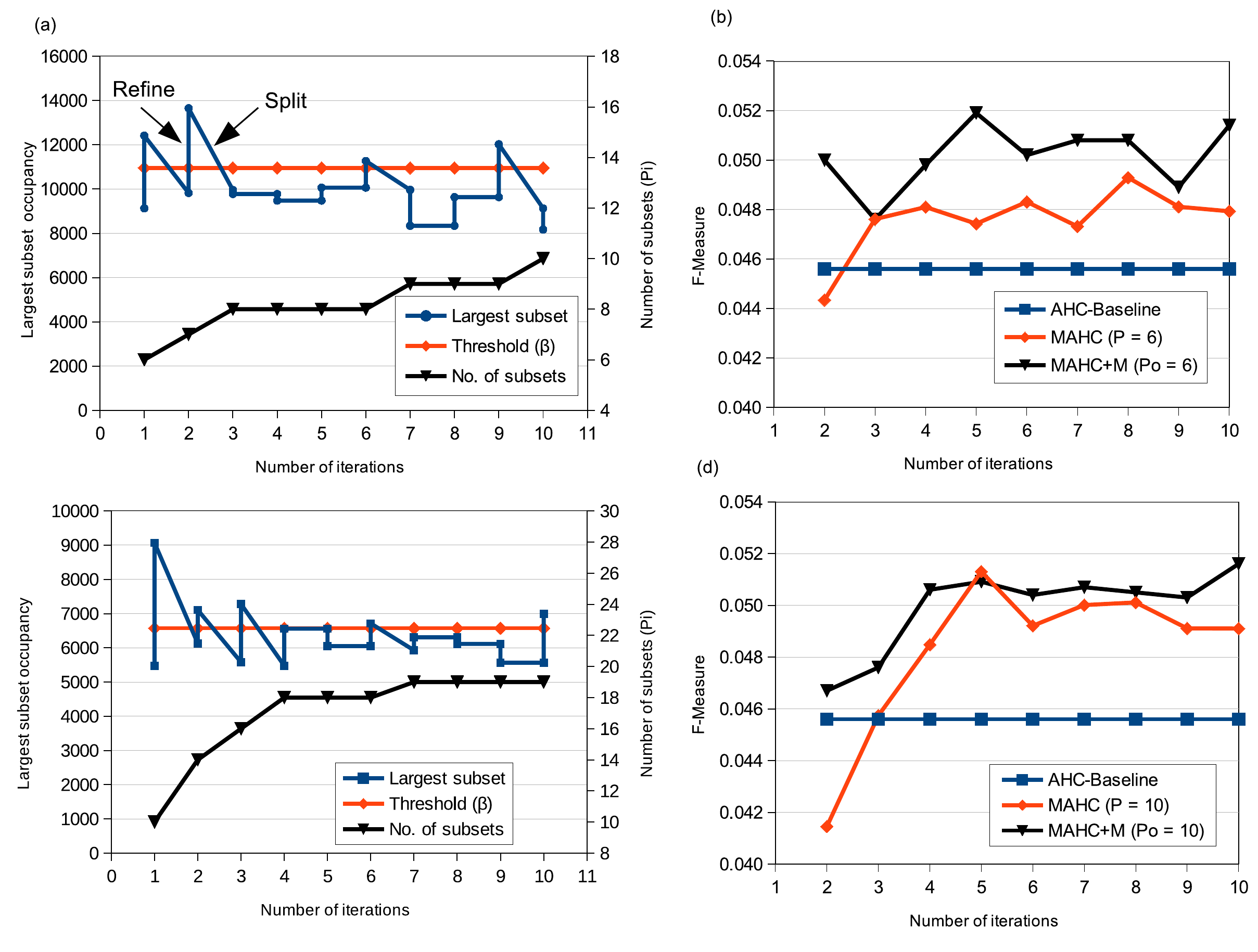}
\caption{Number of subsets  $P_i$ as well as F-measure for each iteration when applying classical agglomerative hierarchical clustering (AHC), modified AHC (MAHC) and MAHC  with cluster size management (MAHC+M) to the Medium Set with an initial number of subsets of $P_0=6$ (a and b)  and $P_0=10$ (c and d).}
\label{fig:mediumcomb}
\end{figure}

Next, we present results for the Medium Set, in this case also explicitly observing the occupancy of the largest subset.
Figure~\ref{fig:mediumcomb} shows the number of subsets $P_i$, the occupancy of the largest subset, and the F-measure when clustering the Medium Set with  $P_0=6$ and  $P_0=10$ initial subsets, as well as example points at which the \emph{split} and \emph{refine} steps in Algorithm~\ref{algo:MAHC+M} occur.

During the \emph{refine} stage, clusters from previous iterations are regrouped, which may lead to greater imbalance in the membership of the clusters and hence an increase in the size of the largest subset. 
The \emph{split} stage subdivides any overly large subsets, ensuring that the threshold $\beta$ is not exceeded. 

Consider for illustration Figure~\ref{fig:mediumcomb}(a), where each subset is occupied by 9,131 segments at the start of the first iteration.  
We observe that the first 2 iterations lead to maximum occupancies that are higher than $\beta$.
In each case the split step subsequently brings these occupancies below the threshold $\beta$.
This also leads to an increase in the number of subsets $P_i$.
Similar behaviour is seen in Figure~\ref{fig:mediumcomb}(c).

Figures~\ref{fig:mediumcomb}(b) and (d) show that, as for Small Sets A and B, the introduction of cluster size management has not led to a degradation in clustering performance in terms of F-measure for the Medium Set.

 
\begin{figure}[!h]
\centering
\includegraphics[scale=0.5]{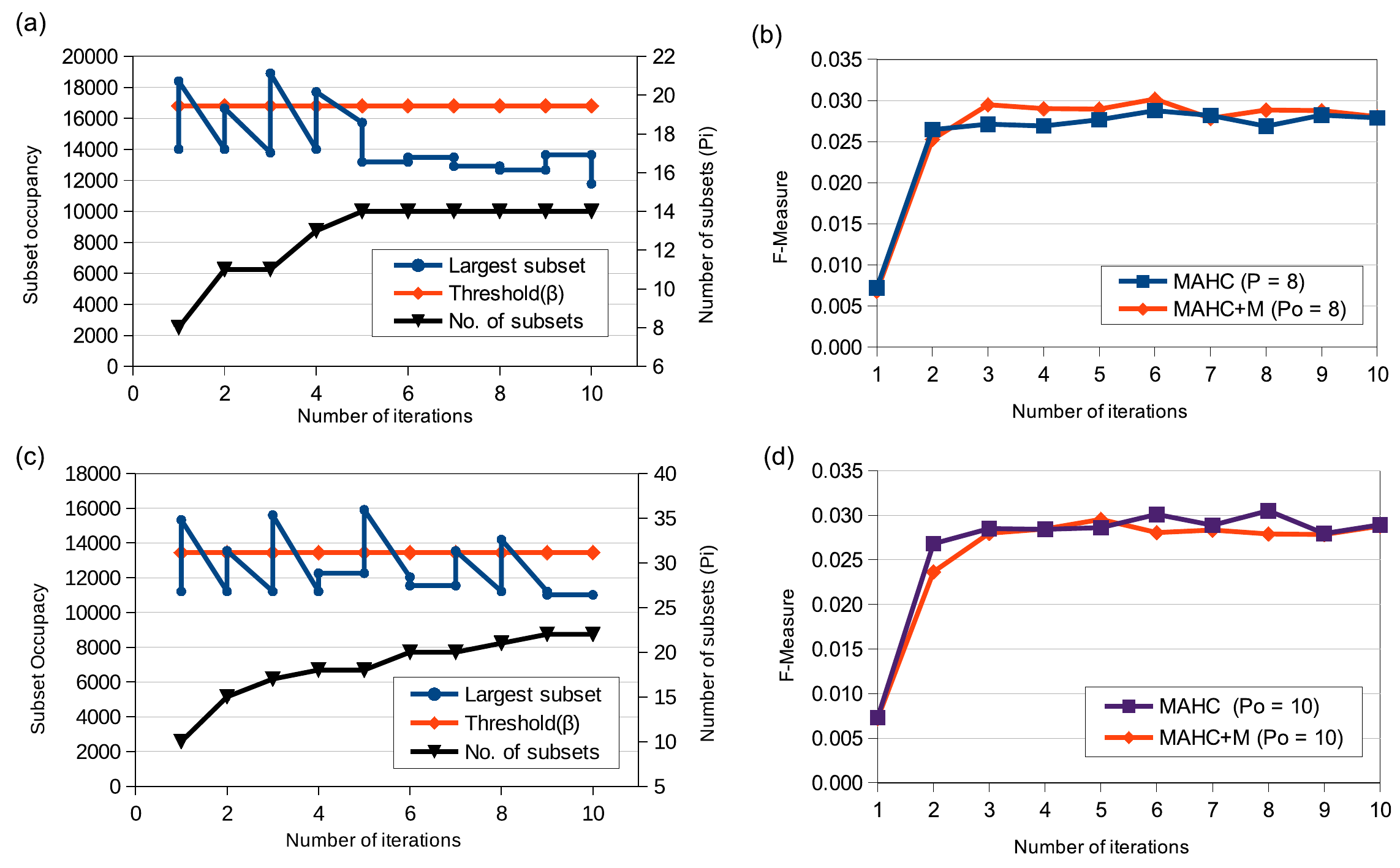}
\caption{Number of subsets $P_i$ as well as F-measure for each iteration when applying modified agglomerative hierarchical clustering (MAHC) and MAHC  with cluster size management (MAHC+M) to the Large Set with an initial number of subsets of $P_0=8$ (a and b)  and $P_0=10$ (c and d).}
\label{fig:largecomb}
\end{figure}

\begin{figure}[!h]
\centering
\includegraphics[scale=0.5]{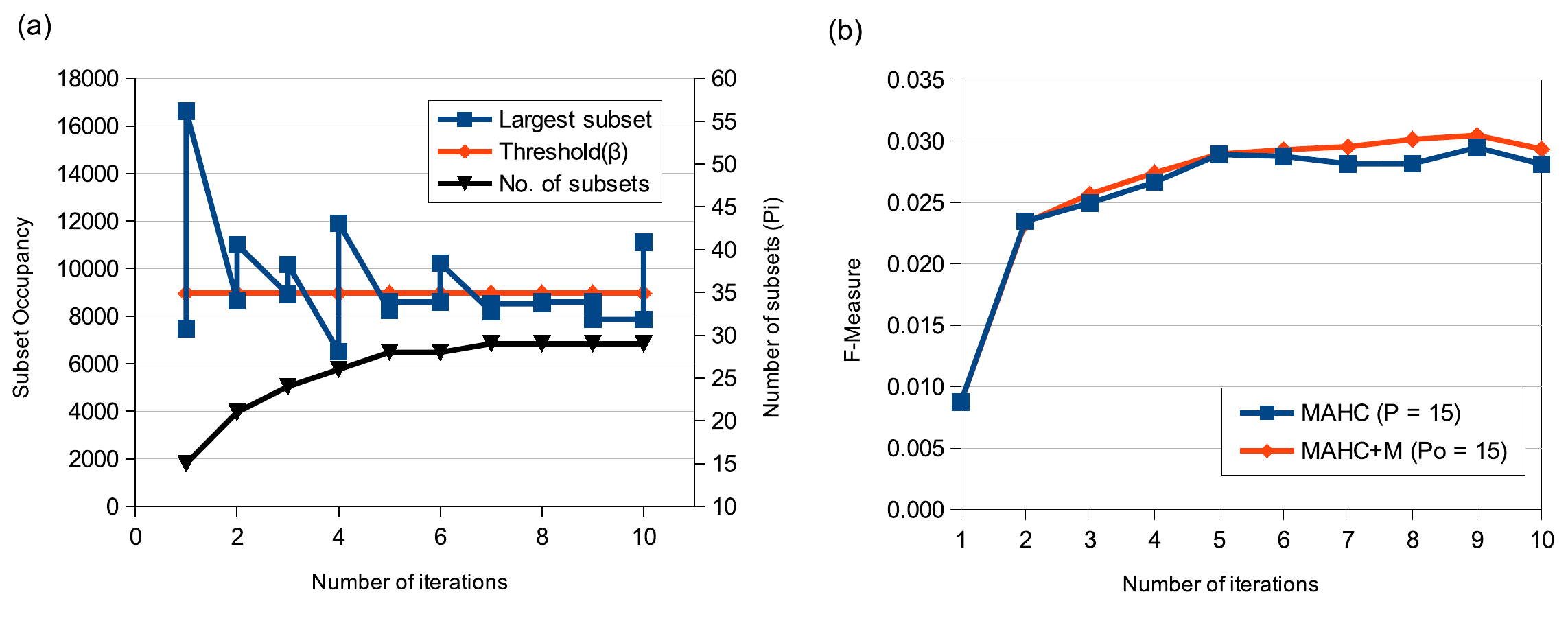}
\caption{Number of subsets $P_i$ as well as F-measure for each iteration when applying modified agglomerative hierarchical clustering (MAHC) and MAHC  with cluster size management (MAHC+M) to the Large Set with an initial number of subsets of $P_0=15$ (a and b).}
\label{fig:subsets15}
\end{figure}

Finally, results for the Large Set are presented in Figure~\ref{fig:largecomb}.
The number of subsets in Figure~\ref{fig:largecomb}(a) where $P_0=8$ reaches a plateau at the fifth iteration, while in  Figure~\ref{fig:largecomb}(c) we see that the number of subsets is still increasing after 8 iterations.
In both cases, however, the corresponding F-measure has settled after 3 iterations, indicating that good clustering has been achieved.  In terms of the F-measure the results for MAHC and MAHC+M are relatively stable and very close in comparison.  
Figure~\ref{fig:subsets15} investigates what happens when the number of subsets is further increased to $P_0=15$. It is observed in Figure~\ref{fig:subsets15} (a) that the number of subsets remain constant from the seventh iteration onwards. The F-measure in Figure~\ref{fig:subsets15} (b) keeps in creasing for both MAHC and MAHC+M cases but settles after the fifth iteration.



\begin{figure}[!ht]
\centering
\includegraphics[scale=0.5]{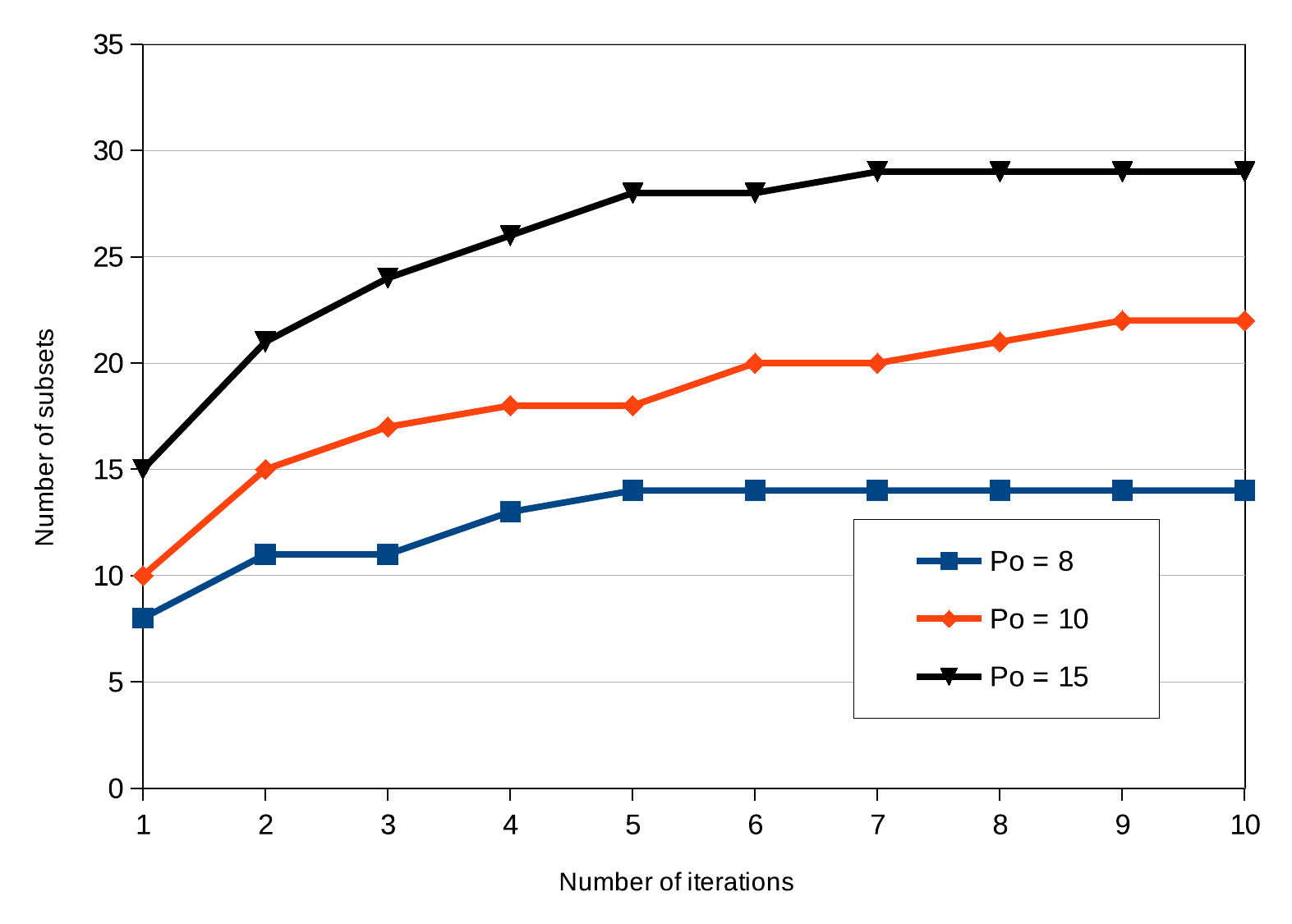}
\caption{Number of subsets  ($P_i$) for each iteration where $P_0$ is initial number of subsets. }
\label{fig:Large}
\end{figure}

Figure \ref{fig:Large} shows how the split step increases the number of subsets used by the clustering algorithm when applied to the Large Set.
We see that the number of subsets seems to settle as the iterations progress.
Referring back to Figure~\ref{fig:largecomb}, we are reminded that the F-measure settles even when the number of subsets continues to increase.
Hence it appears to be reasonable to terminate the clustering algorithm after a fixed number of iteration, and it is not necessary to wait for example until the number of subsets no longer changes.

\begin{figure}[!h]
\centering
\includegraphics[scale=0.5]{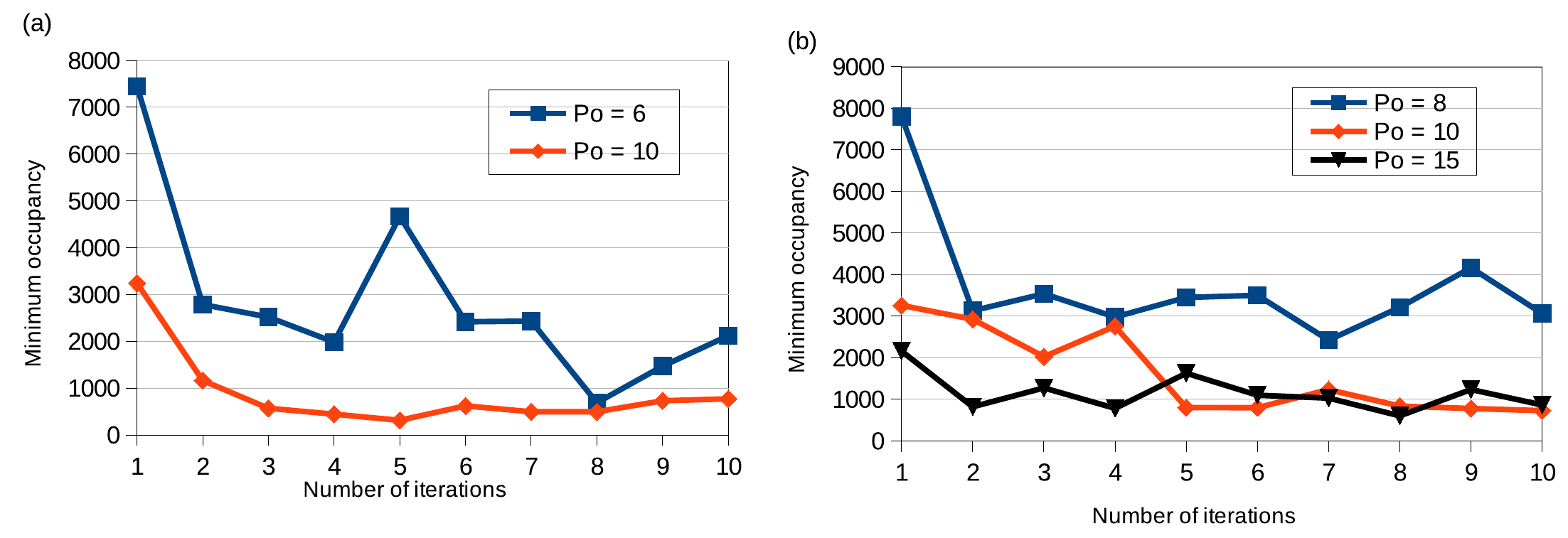}
\caption{Minimum occupancy per iteration for (a) Medium Set and (b) Large Set.}
\label{fig:minoccupancy}
\end{figure}

Finally, we would like to consider the merit of introducing a \textit{merge} step to complement the \textit{split} step into Algorithm~\ref{algo:MAHC+M}.
The motivation for a merge step would be to re-absorb subsets whose membership vanishes during the algorithm due to the repeated iterative application of the split step.
Figure~\ref{fig:minoccupancy} shows the size of the smallest subset at each iteration for the Medium and Large Sets.
We see that for both datasets the subset membership never vanishes.
From this we conclude that the addition of a merge step is not necessary for the effective functioning of the algorithm.


\section{Summary and conclusion}

Modified agglomerative hierarchical clustering (MAHC) has recently been presented as a divide-and-conquer strategy allowing agglomerative hierarchical clustering (AHC) to be applied to otherwise impractically large datasets.
We have extended this algorithm by iteratively enforcing a hard limit on the size of the data subsets processed by MAHC.
This allows maximum space constraints to be guaranteed, and makes MAHC more reliably useful for the hierarchical agglomerative clustering of large datasets.
We show that the proposed modification does not affect the algorithms performance in terms of F-measure when applied to a number of datasets of varying size compiled from the TIMIT speech corpus.
in ongoing work, the algorithm is being applied to the automatic induction of subword units and associated pronunciations for use in automatic speech recognition (ASR).


\end{document}